\documentclass[10pt,twocolumn,letterpaper]{article}

\usepackage{iccv}              

\usepackage[table]{xcolor}
\usepackage{colortbl}
\usepackage{multirow}
\usepackage{lineno}

\definecolor{iccvblue}{rgb}{0.21,0.49,0.74}
\usepackage[pagebackref,breaklinks,colorlinks,allcolors=iccvblue]{hyperref}

\def\paperID{8785} 
\def\confName{ICCV}
\def\confYear{2025}

\title{DuoCLR: Dual-Surrogate Contrastive Learning 

for Skeleton-based Human Action Segmentation 
}

\author{Haitao Tian, Pierre Payeur\\
University of Ottawa, Canada\\
{\tt\small \{htian026, ppayeur\}@uottawa.ca} %
}

\begin{document}
\maketitle
\begin{abstract}
In this paper, a new contrastive representation learning framework is proposed to enhance action segmentation via pretraining using trimmed (single action) skeleton sequences. Unlike previous representation learning works that are tailored for action recognition and that develop isolated sequence-wise representations, the proposed framework focuses on exploiting multi-scale representations in conjunction with cross-sequence variations. More specifically, it proposes a novel data augmentation strategy, “Shuffle and Warp”, which exploits diverse multi-action permutations. The latter effectively assist two surrogate tasks that are introduced in contrastive learning: Cross Permutation Contrasting (CPC) and Relative Order Reasoning (ROR). In optimization, CPC learns intra-class similarities by contrasting representations of the same action class across different permutations, while ROR reasons about inter-class contexts by predicting relative mapping between two permutations. Together, these tasks enable a Dual-Surrogate Contrastive Learning (DuoCLR) network to learn multi-scale feature representations optimized for action segmentation. In experiments, DuoCLR is pre-trained on a trimmed skeleton dataset and evaluated on an untrimmed dataset where it demonstrates a significant boost over state-the-art comparatives in both multi-class and multi-label action segmentation tasks. Lastly, ablation studies are conducted to evaluate the effectiveness of each component of the proposed approach.
\end{abstract}    
\section{Introduction}
\label{sec:intro}

Contrastive learning \cite{guo2022contrastive, chen2020simple, chen2020improved} opens a new opportunity for taking advantage of the growing volume of unlabeled data for learning a rich latent feature representation aware of underlying data structures and semantics without human annotations. This approach effectively reduces the need for extensive labeled data in downstream task fine-tuning. In human-centric video understanding, human skeleton data \cite{das2019toyota, liu2017pku, shahroudy2016ntu} has become a widely used modality in contrastive learning as it depicts the trajectories of key body joints and provides a compact and robust depiction of human motion that is resilient to environmental variations. Existing research primarily involves trimmed (single action) skeleton sequences into contrastive learning, where the model treats each sequence as an instance and learns intra-action temporal representations (i.e., \textit{similarities}) through instance discrimination \cite{chen2021channel, chi2024infogcn++, yan2018spatial, yang2021unik}. The learned representations have been effectively transferred into trimmed-sequence-based downstream tasks such as \textit{action recognition} \cite{chi2024infogcn++, abdelfattah2024maskclr, franco2023hyperbolic, huang2023graph, lin2023actionlet, thoker2021skeleton, wu2024scd,yang2023lac, zhu2023modeling} and \textit{action retrieval} \cite{guo2022contrastive,dong2023hierarchical, shah2023halp,Zhang2023, li20213d}.

\begin{figure}
    \centering
    \includegraphics[width=1\linewidth]{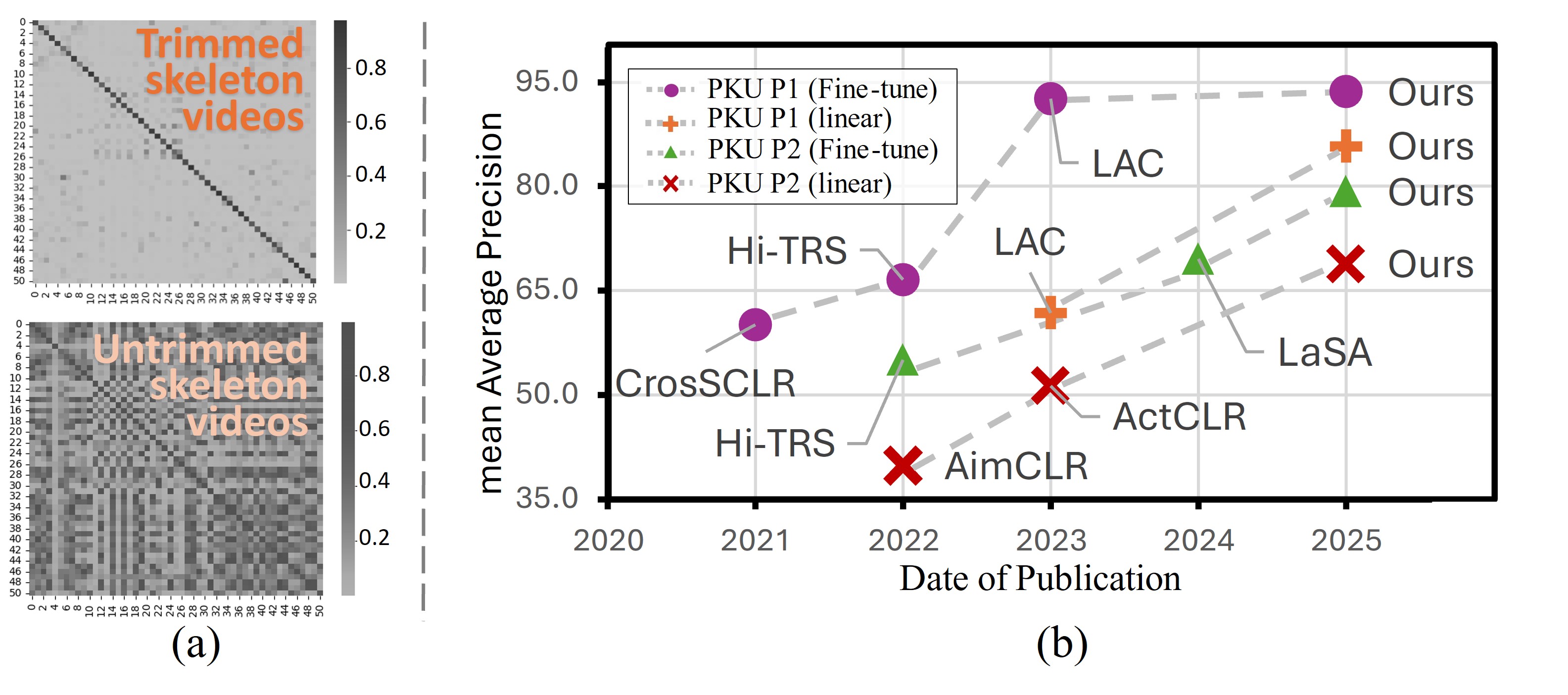}
    \vspace{-20pt}
    \caption{Limitation of traditional skeleton-based contrastive learning. (a): Distinct feature similarity matrices obtained from feature extraction on trimmed (upper) vs. untrimmed (lower) skeleton videos. Entry (i, j) represents the feature similarity (Eq. (1)) between action classes i and j. The feature extractor is learned on NTU RGB+D using AimCLR \cite{guo2022contrastive}. Trimmed and untrimmed videos are both sampled from PKU Part 1. For untrimmed videos, action-specific features are extracted via post-segmentation in the feature space using frame annotations (detailed in Sec. 3.3). (b): Trend of contrastive learning performance (detailed in Table \ref{table1}) in skeleton action segmentation by publication year.}
    \label{fig1}
\end{figure}

However, complex human-centric video understanding applications such as \textit{action segmentation} often involve untrimmed long skeleton videos exhibiting multiple actions and complex action correlations. While focusing on sequence-wise instance discrimination tasks, traditional contrastive learning generally neglects the exploitation of inter-sequence temporal dependencies (i.e., \textit{contexts}) that are critical for action segmentation. To highlight this limitation, Fig. \ref{fig1}a experimentally verifies that a state-of-the-art contrastive learning paradigm \cite{guo2022contrastive}, which learns isolated sequence-wise representations, may lead to significantly different feature representations when applied to trimmed and untrimmed videos. We hypothesize that such a discrepancy results from the extraction of inconsistent action representations from dynamic action contexts permuted in untrimmed sequences, which leads to suboptimal transfer learning performance  to action segmentation tasks (as illustrated in Fig. \ref{fig1}b).

In this work, a new contrastive learning framework is proposed to improve action segmentation on \textit{untrimmed} videos by pretraining on \textit{trimmed} skeleton videos. \textbf{First, the framework}\textbf{ leverages }\textbf{multi-action }\textbf{permutations to fulfill cross-sequence data augmentation. }Previous single-action augmentations use simple sequence-wise modifications, e.g., spatial transformation \cite{guo2022contrastive,chi2024infogcn++} and temporal scaling \cite{li2019actional, Zhang2023}, that generate low-level context-free data variations. This work rather introduces “Shuffle and Warp” to exploit high-level contextual variations: (1) “Shuffle” treats trimmed skeleton sequences as modular action elements that can be sampled and shuffled, introducing dynamic cross-sequence variations; and (2) “Warp” leverages geometrical transformations to concatenate the shuffled sequences into a view-consistent permutation of actions.

\begin{figure}
    \centering
    \includegraphics[width=1\linewidth]{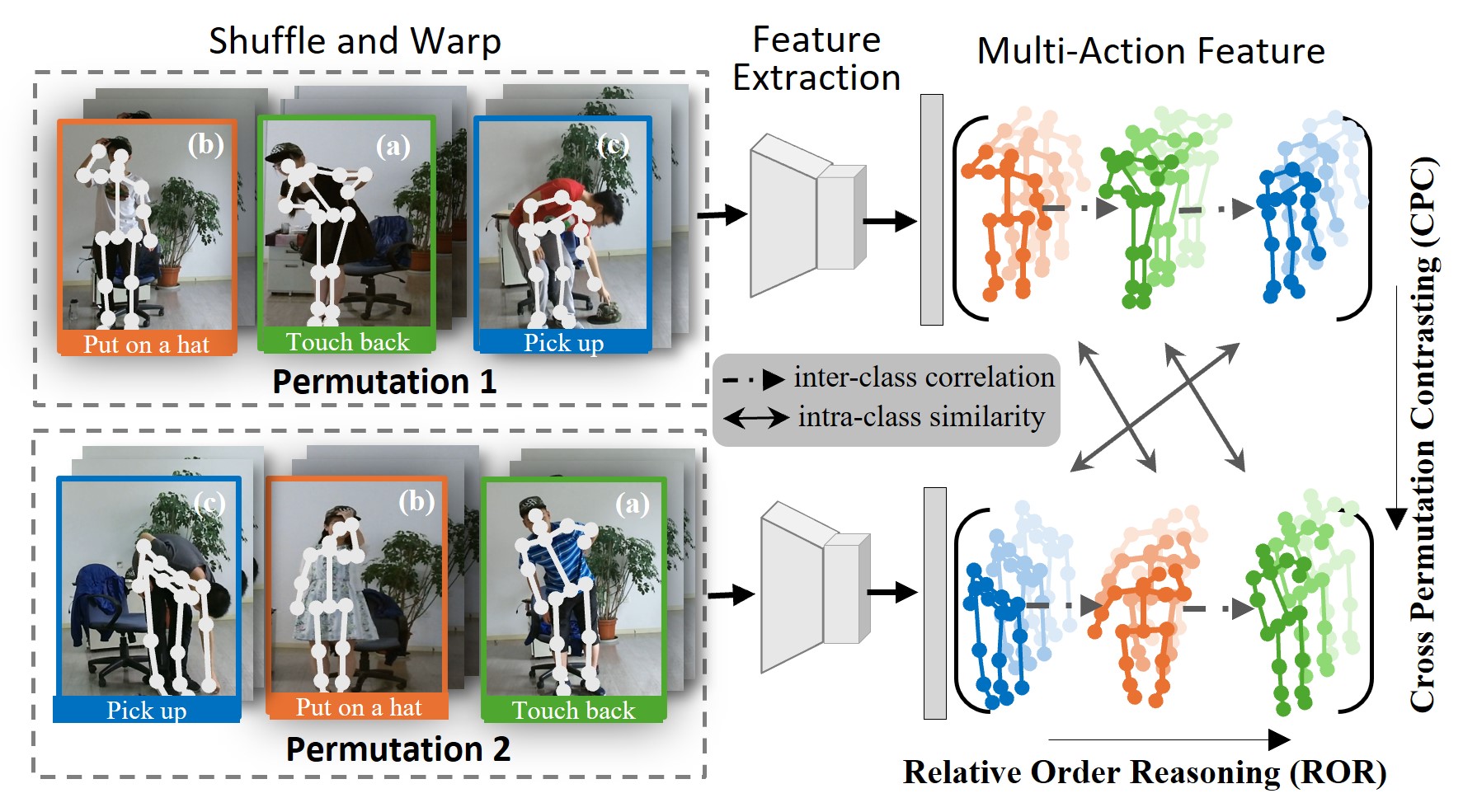}
    \vspace{-20pt}
    \caption{The proposed contrastive learning framework for action segmentation utilizes multi-action permutations. For example, six single-action skeleton sequences are randomly shuffled and warped into two sequences, Permutation 1 and Permutation 2, with three shared actions. Such a data augmentation introduces dynamic cross-sequence variations that enhance learning in two ways: i) CPC contrasts intra-class instance across different permutations, promoting permutation-invariant representations despite changing contexts (e.g., "Put on a hat" appears in both permutations but in different contexts); ii) ROR  captures inter-class dependencies by mapping one permutation to another (e.g., learning the transformation from (b, a, c) to (c, b, a)). }
    \label{fig2}
\end{figure}

\textbf{Second, two new surrogate tasks ensure }\textbf{multi-scale representation learning.} To fully exploit the temporal variations present in multi-action permutations, this work introduces two contrastive learning surrogate tasks, Cross Permutation Contrasting (CPC) and Relative Order Reasoning (ROR). As conceptually illustrated in Fig. \ref{fig2}, CPC learns \textit{permutation}\textit{-invariant} \textit{intra-class }\textit{similarities} via instance discrimination where the representation of the same action class should be similar regardless of context variations arising from their neighboring segments. ROR focuses on the \textit{permutation-aware} \textit{inter-class} \textit{contexts} by predicting the temporal mapping between action permutations, encouraging the model to learn relative positioning among actions, thereby adding a complementary regularization to the feature space. 

\textbf{Lastly, a hierarchical network facilitates sliding-window-free transfer learning.} We amalgamate the key heuristics into a \textbf{Du}al-Surr\textbf{o}gate \textbf{C}ontrastive \textbf{L}ea\textbf{R}ning (DuoCLR) network that can be trained end-to-end. Unlike previous works \cite{yang2023lac, chen2022hierarchically, lin2023actionlet, zhang2023prompted} relying on a single visual encoder, DuoCLR employs a cascaded structure \cite{filtjens2022skeleton} composed of a visual encoder and a temporal encoder. The additional temporal encoder enables a large temporal receptive field, allowing the model to capture long-range action dependencies without requiring sliding-window operations \cite{yang2023lac, dai2022toyota, chen2022hierarchically} on untrimmed skeleton videos. As a result, DuoCLR surpasses state-of-the-art methods in action segmentation (Fig. \ref{fig1}b). After pretraining, both encoders can be jointly fine-tuned as a feature extractor for downstream tasks.

The original contributions of the work are threefold. First, a new contrastive learning framework for skeleton data-based representation learning enables trimmed sequences to learn a fine-grained representation for action segmentation. Second, a new data augmentation strategy, "Shuffle and Warp", for skeleton data introduces variations in between data sequences to improve feature learning. Third, the framework builds upon a dual-contrastive approach by combining CPC and ROR. It focuses on instance coherence and instance correlation, both critical for action segmentation. The project is available at: \href{https://htian026.github.io/DuoCLR}{\texttt{https://htian026.github.io/DuoCLR}}, which contains code implementation.

\section{Related work}
\label{sec:realtedwork}
\textbf{Skeleton-based Action Segmentation.} Compared to skeleton action recognition \cite{chen2021channel, yan2018spatial, yang2021unik, li2019actional, duan2023skeletr, xiang2023generative} that aims at sequence-wise action classification, skeleton action segmentation \cite{chai2024motion, chen2020action,filtjens2022skeleton, li2023decoupled,ma2021fine} is inherently more challenging as the model must aggregate both action patterns and contexts among multiple actions permutated in a single sequence. To tackle the difficulty, recent research \cite{ji2025language,filtjens2022skeleton,li2023decoupled, ji2025language} proposes to optimize an end-to-end model that incorporates frame-wise action classification into supervised training. Yet, training an action segmentation model from scratch relies heavily on frame-level annotations that are costly and time-consuming to generate \cite{liu2017pku, das2019toyota}. In this work, we circumvent the data annotation usage by leveraging trimmed skeleton videos for pretraining, where a contrastive learning network can leverage numerous action permutations to pre-learn a feature extractor relevant to action segmentation, and quick fine-tuning to untrimmed videos.

\textbf{Skeleton Data Augmentation} is primarily achieved by single-action skeleton modifications. For instance, studies \cite{guo2022contrastive, abdelfattah2024maskclr, lin2023actionlet, thoker2021skeleton, wu2024scd, li20213d, huang2023graph, Zhang2023, chen2022hierarchically, zhou2023self} delve into designing more effective single-action augmentations, e.g., “Shear”, “Crop”, “Spatial Flip”, “Spatial Rotate”, “Spatial Jitter”, “Axis Mask”, “Gaussian Noise”, and “Gaussian Blur”, demonstrating the effectiveness of carefully involving multiple augmentations in representation learning \cite{guo2022contrastive, Zhang2023}. Recent works also leverage multi-action skeleton augmentations, e.g., “Latent Action Composition” \cite{yang2023lac} and “Skeleton Actionlet” \cite{lin2023actionlet} are proposed to mix spatiotemporal statistics of two sequences; “Skeleton-Mix” \cite{lin2023actionlet, chen2022contrastive, hu2024global, liu2023skeleton, xiang2024joint, zhang2024shap} is proposed to generate a synthetic sequence by merging both data and label from two sequences into a composable sequence. 
SCS \cite{tian2025stitch} proposes a temporal skeleton stitching scheme to generate multi-action stitched sequences. However, it imposes an offline data generation process requiring time-consuming frame registration operations. In contrast, this work proposes utilizing online multi-action permutations, "Shuffle and Warp", a new data augmentation scheme which will be detailed in Section 3.2.

\textbf{Skeleton-based Representation Learning} \cite{chen2020simple,chen2020improved}  is primarily achieved by learning action similarities from augmented skeleton views via effective surrogate tasks such as instance discrimination \cite{guo2022contrastive, chi2024infogcn++, abdelfattah2024maskclr,dong2023hierarchical,franco2023hyperbolic, huang2023graph, lin2023actionlet,  wu2024scd, zhu2023modeling, li20213d, Zhang2023, chen2022hierarchically, zhou2023self,  chen2022contrastive, hu2024global} , jigsaw puzzles \cite{lin2020ms2l, su2021self}, adversarial learning \cite{zheng2018unsupervised}, and language-video contrasting \cite{ji2025language}. However, prior works mainly produce sequence-level representations, which are effective for simplified downstream tasks like action recognition but may be insufficient for action segmentation, where fine-grained feature representations are crucial. Recent efforts  \cite{dai2022toyota, yang2023lac, chen2022hierarchically} attempt to mitigate this by applying sliding windows to generate fixed-length action clips from untrimmed sequences, enabling the use of standard action recognition models. However, this strategy introduces noisy boundaries, leading to degraded segmentation accuracy (experimentally verified in Section 4.3). These limitations highlight the need for a contrastive learning paradigm tailored for action segmentation.

\section{Dual-Surrogate Contrastive Learning}
\subsection{Overview}

\begin{figure*}[t]
    \centering
    
    \includegraphics[width=1\linewidth]{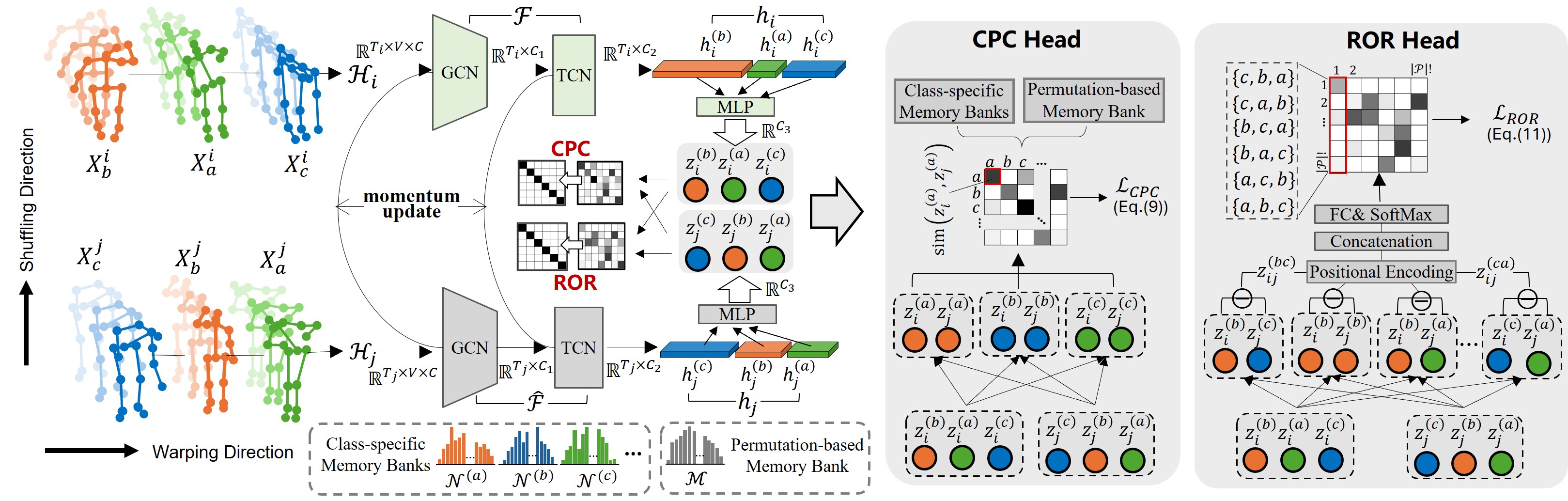}
   \vspace{-20pt}
    \caption{Computational   workflow of the proposed DuoCLR approach. Trimmed single-action skeleton videos are augmented by using “Shuffle and Warp” on the input. In DuoCLR, the two surrogate tasks, “ROR” and “CPC”, share the feature encodings, \(z_i^{(n)}\)and \(z_j^{(n)}\), while fulfilling different pipelines for representation learning.  By solving simultaneous CPC and ROR, the network learns a fine-grained action representation for action segmentation.}
    \label{fig3}
\end{figure*}

Let \({X\in\mathbb{R}}^{T\times V\times C}\) be a trimmed skeleton video with \textit{T} skeleton frames \textit{V} skeleton joints, and \textit{C} spatial dimensions, which covers the motion of a single action class. A contrastive learning paradigm aims to use trimmed skeleton videos to pretrain a feature extraction network \(\mathcal{F}\) by which the features of the same action class should be similar. In previous studies \cite{guo2022contrastive, huang2023graph, li20213d}, feature similarity between two sequences, \(X_i\) and\( X_j\), is calculated as: 
\begin{equation}
\small
    \begin{aligned}
        \text{sim}\big(\mathcal{F}(X_i), \mathcal{F}(X_j)\big) &= \frac{\mathcal{F}(X_i) \ \mathcal{F}(X_j)}{\lvert \mathcal{F}(X_i) \rvert \lvert \mathcal{F}(X_j) \rvert\cdot\tau}
    \end{aligned}
\end{equation}
where \(\tau\) acts as the temperature parameter.
This work aims to pretrain a \(\mathcal{F}\) using Dual-Surrogate Contrastive Learning (DuoCLR) and customize a task-specific (action segmentation) layer \(\mathcal{C}\) for a target dataset composed of a small number of frame-wise labeled and untrimmed videos. Section 3.2 describes the Shuffle and Warp data augmentation technique for generating dynamic multi-action permutations. Section 3.3 details the network structure of \(\mathcal{F}\). In Sections 3.4 and 3.5, DuoCLR introduces two surrogate tasks (as conceptually depicted in Fig. \ref{fig3}), Cross Permutation Contrasting (CPC) and Relative Order Reasoning (ROR), respectively. 

\subsection{Multi-Action Permutations}

Capturing temporal contexts is critical for action segmentation as it requires the model to discern the state of actions over a single untrimmed skeleton sequence. Since trimmed skeletal data lacks such context, Shuffle and Warp is introduced to generate dynamic multi-action permutations to support improved contrastive learning for action segmentation.

\textbf{Temporal Shuffling} was previously utilized to generate single-action augmentation where segments of a trimmed sequence are shuffled to solve a jigsaw puzzle task  for action recognition \cite{lin2020ms2l,Misra2016}. In this work, we rather involve multiple trimmed sequences at each augmentation of temporal shuffling to introduce cross-sequence variations. Specifically, let \( \mathcal{X}=\{X^{(a)},X^{(b)},X^{(c)},\ldots \}\) be a set of trimmed skeleton sequences sampled from a data batch, we can generate a multi-action permutation as:
\begin{equation}
    \begin{aligned}
        \text{\textit{Shuffle}}(\mathcal{X}, \mathcal{P}) = \text{\textit{concat}} \left( X^{\mathcal{P}[1]}, X^{\mathcal{P}[2]}, X^{\mathcal{P}[3]}, \dots \right)
    \end{aligned}
\end{equation}
where the function “\textit{Shuffle}” first permutes the elements of \(\mathcal{X}\) according to the permutation \(\mathcal{P}\), e.g.,\(\mathcal{P}=(b,c,a,...)\), and then generates a single sequence using the tensor concatenation operation “\textit{concat}”. The proposed data augmentation supports both labeled and unlabeled skeleton data where \(a,b,c,...\) denotes action classes while labeled and sequence indices while unlabeled (discussed in Section 4.3). 

\textbf{Temporal Warping.} However, skeleton sequences collected under different imaging configurations may lead to unnecessary data variations in the skeleton scale and poses \cite{shahroudy2016ntu}. A representation learning network may use such intricacies to develop shortcuts to fulfill the surrogate task of contrastive learning \cite{guo2022contrastive, thoker2021skeleton}. To tackle the issue, temporal warping is newly proposed to eliminate the unnecessary data variations by using a pairwise sequence concatenation operation. Specifically, given two heterogenous sequences \(X_i\in\mathbb{R}^{T_i\times V\times C}\) and \(X_j\in\mathbb{R}^{T_j\times V\times C}\), it uses a rotation \(\mathcal{R}{\in\mathbb{R}}^{C\times C}\),  a translation \(\mathcal{T}{\in\mathbb{R}}^{C\times1}\) , and a scaling factor \(s\in\mathbb{R}\) , to map \(X_j\) into the camera configuration of \(X_i\). In this way, “\(Warp\)” is defined as: 
\begin{equation}
    \begin{aligned}
         Warp\left(X_i,X_j\right)=concat\left(X_i,s\left(X_j\ast\mathcal{R}+\mathcal{T}\right)\right) 
    \end{aligned}
\end{equation}
where \(\ast\) denotes the batched dot product. Details regarding parameters estimation and sequence operation are available in supplementary material.   

At last, we use “Shuffle and Warp” to generate the multi-action permutation \( \mathcal{H}\) as follows
\begin{equation}
    \begin{aligned}
\mathcal{H}=Warp(Shuffle(\mathcal{X},\mathcal{P})),    
    \end{aligned}
\end{equation}
which denotes that \(\mathcal{H}\) is generated by recursively operating the “\textit{Warp}” operation on shuffled \(\mathcal{X}\) under \(\mathcal{P}\).

\subsection{Feature Extraction}

In this work, the feature extraction network \(\mathcal{F}\) is composed of a visual encoder and a temporal encoder, as illustrated in Fig. \ref{fig3}.

\textbf{Visual Encoder.} It is common in the research community \cite{dong2023hierarchical, franco2023hyperbolic, lin2023actionlet, huang2023graph, li2019actional, lin2023actionlet} to use a graph convolutional network (GCN) \cite{yan2018spatial} as a visual encoder to parse intermediate skeleton feature embeddings. In this work, we removed the temporal pooling layer of GCN to keep the feature’s temporal resolution, i.e., GCN:\(\mathbb{R}^{T\times V\times C}\rightarrow\mathbb{R}^{T\times C_1}\). However, a vanilla GCN may be limited by their small temporal receptive fields \cite{filtjens2022skeleton}, thus not exploiting contextual structures that are relevant to action segmentation. 

\textbf{Temporal Encoder.} In order to effectively capture action relationships, we stack a temporal encoder onto the visual encoder to constructs  \(\mathcal{F}\) . The temporal encoder adopts a temporal convolution network \cite{farha2019ms} that utilizes a series of dilated temporal convolutions to aggregate long-range temporal contexts of the hidden state from the visual encoder, i.e., \(TCN:\mathbb{R}^{T\times C_1}\rightarrow\mathbb{R}^{T\times C_2}\).

At last, given a skeleton sequence input \(X\), its hidden feature representation \(h\) is obtained as:
\vspace{-2pt}
\begin{equation}
\small
    \begin{aligned}
           h = \underbrace{(\text{TCN} \circ \text{GCN})}_{\mathcal{F}}(X)
 \end{aligned}
\end{equation}
\vspace{-10pt}

\textbf{Multi-grained feature projections. }DuoCLR differs from prior contrastive learning works that input single-action sequences. DuoCLR rather utilizes multi-action permutations for feature extraction prior to contrastive learning. As illustrated in Fig. \ref{fig3}, DuoCLR uses Shuffle and Warp (Eq. (4)) to generate a pair of multi-action permutations \(\mathcal{H}_i\) and \(\mathcal{H}_j\) of \(\mathcal{X}\) with two random permutations \(\mathcal{P}_i\) and \(\mathcal{P}_j\). Using Eq. (5), it obtains multi-action representations \( h_i=\mathcal{F}(\mathcal{H}_i\ ) \) and \( h_j=\bar{\mathcal{F}}(\mathcal{H}_j)\), where \(\bar{\mathcal{F}}\) denotes a moving average of the feature extractor  \(\mathcal{F}\) and we follow the same implementation introduced in \cite{guo2022contrastive}. For notation simplicity, let’s assume the cardinality of \(\mathcal{X}\) is 3, e.g., \(\mathcal{X}={X^{\left(a\right)},X^{\left(b\right)},X^{\left(c\right)}}\) , and assume \(\mathcal{P}_i=(b,a,c) \) and \(\mathcal{P}_j=(c,b,a)\). In this way, DuoCLR generates multi-grained feature projections for contrastive learning by treating the feature representations, \(h_i\) and \(h_j\), as concatenations of action-wise encodings:   \(h_i=concat(h_i^{\left(b\right)},h_i^{\left(a\right)},h_i^{\left(c\right)})\) and  \( h_j=concat\left(h_j^{\left(c\right)},h_j^{\left(b\right)},h_j^{\left(a\right)}\right)\). Furthermore, an MLP-based projection head is used to generates \textit{local} and \textit{global} feature projections:
\begin{equation}
    \begin{aligned}
         z_i^{\left(n\right)}=MLP[\phi(h_i^{\left(n\right)})]; z_i=MLP[\phi(h_i)]
     \end{aligned}
\end{equation}
where n=a,b,c and \( \phi\)\ denotes the temporal average pooling operation. Likewise, it generates  \(z_j^{\left(n\right)}\) and \(z_j\).

\subsection{Cross Permutation Contrasting (CPC)}

While treating each action within a permutation as an instance, individual instances of the same action class can have different contextual neighbors (e.g., the action "Put on a hat" in Fig. \ref{fig2} may appear next to different actions in each permutation). Traditional methods \cite{yang2023lac, guo2022contrastive, li2019actional, lin2023actionlet} overlook such information and learn isolated sequence-wise representations, thereby achieving suboptimal transferability in action segmentation (as illustrated in Fig. \ref{fig1}). In this work, DuoCLR utilizes multi-action permutations as data augmentation and employs Cross Permutation Contrasting (CPC) for self-supervised representation learning. Since the augmentation of action permutations expose the network to numerous action temporal contexts, CPC learns permutation-invariant temporal coherence where intra-class representations from the different permutations should be pulled together while inter-class representations should be pushed away. 

\textbf{Permutation Bank-based Instance Discrimination.} In implementation, CPC employs local feature projections \(z_i^{\left(n\right)}\in\mathbb{R}^{C_3}\) and \( z_j^{\left(n\right)}\in\mathbb{R}^{C_3}\) as positive pairs and formulates class-specific memory banks \( \mathcal{N}=\{{\mathcal{N}^{\left(m\right)}}\}\) to store negative instances, each updated with \(z_j^{\left(m\right)}\) where \(m\neq\ n\). Afterwards, it learns short-term invariant representations for each instance regardless of surrounding context changes by optimizing a InfoNCE loss \cite{chen2020improved}:
\begin{equation}
\small
    \begin{aligned}
        \mathcal{L}_\mathcal{N}^{\left(n\right)}=-log\frac{sim{\left(z_i^{\left(n\right)},z_j^{\left(n\right)}\ \right)}}{sim{\left(z_i^{\left(n\right)},z_j^{\left(n\right)}\ \right)}+\sum_{z^-\in\mathcal{N}}sim{\left(z_i^{\left(n\right)},z^{-}\right)}} 
     \end{aligned}
\end{equation}
where \( z^-\) is a negative sample from the memory bank.

Second, CPC also involves a permutation (multi-action) memory bank \(\mathcal{M}\) in contrastive learning, which is updated by \(z_j\).  When \( \mathcal{H}_i \) and \(\mathcal{H}_j\) share the same permutation, i.e., \(\mathcal{P}_i=\mathcal{P}_j\), it treats each permutation as an instance and encourages the network to learn long-term invariant representations of permutations. It calculates the similarity between permutation-wise positive pair, \(z_i\in\mathbb{R}^{C_3}\) and\( z_j\in\mathbb{R}^{C_3}\), and optimizes the second InfoNCE loss as:
\begin{equation}
    \begin{aligned}
        \mathcal{L}_\mathcal{M}=-log\frac{sim(z_i,z_j)}{sim\left(z_i,z_j\right)+\sum_{z^- \in\mathcal{M}}sim{\left(z_i,z^-\right)}}
\end{aligned}
\end{equation}
Upon the formulation of Eq. (7) and Eq. (8), the final objective for CPC is formulated as: 
\begin{equation}
    \begin{aligned}
        \mathcal{L}_{CPC}=\lambda\sum_{n=a,b,c,...}\mathcal{L}_\mathcal{N}^{\left(n\right)}+(1-\lambda)\mathcal{L}_\mathcal{M} 
\end{aligned}
\end{equation}
where \(\lambda\) is equals to 1 if\ \(\mathcal{P}_i\neq\mathcal{P}_j\), and 0 otherwise.

\subsection{Relative Order Reasoning (ROR)}

Order reasoning of action videos is a fundamental task in video representation learning. Traditional approaches primarily focus on chronological order prediction within a single-action class, where models learn temporal coherence by reordering shuffled video frames \cite{Lee2017} or clips \cite{lin2020ms2l, Misra2016, Xu2019}. However, in multi-action sequences, the order is inherently ambiguous, making traditional approaches not applicable. To address this, we introduce Relative   Order Reasoning (ROR), a novel paradigm that leverages multi-action permutations. Instead of predicting a single "correct" order, ROR focuses on “relative” mapping between paired action permutations and learns the capability of discerning the latent state of actions over long skeleton sequences, thus regularizing the feature space from a complementary direction to CPC.

In implementation, DuoCLR solves ROR via classification upon the mapping between \( \mathcal{H}_i\) and \(\mathcal{H}_j\). Specifically, since each pair of action permutations share the same actions, the relative mapping between the paired permutations is uniquely determined. For example, let \(|\mathcal{P}_j|\) (where \( \mathcal{P}_j=\left(c,b,a\right)\)) be the action granularity of \( {\mathcal{H}}_j\), there exist \(|\mathcal{P}_j|\ !=6\) factorial orders: \(\{{(c,b,a), (c,a,b),(b,c,a),(b,a,c)...}\}\). The mapping from \(\mathcal{H}_i\) to \(\mathcal{H}_j\) is represented by the one-hot class label \(\textbf{1}(\mathcal{P}_i|\mathcal{P}_j)=\left[0,0,0,1,0,0\right]^T\) while \(\mathcal{P}_i=(b,a,c)\).

\textbf{Positional Encodings .} For optimization, ROR calculates Positional Encodings (PE) of \( \mathcal{H}_i\) and \(\mathcal{H}_j\). It uses local projections \(z_i^{\left(n\right)}\in\mathbb{R}^{C_3}\) and \( z_j^{\left(m\right)}\in\mathbb{R}^{C_3}\) (obtained via Eq. (6)) to calculate pair-wise feature comparison \(z_{ij}^{\left(n,m\right)}=\left|z_i^{\left(n\right)}-z_j^{\left(m\right)}\right|\). Second, it generates the concatenation of \(\{{z_{ij}^{\left(n,m\right)}|n\in\mathcal{P}_i,m\in\mathcal{P}_j}\}\) to obtain fine-grained position relations between \(\mathcal{H}_i\) and \(\mathcal{H}_j\): 
\begin{equation}
    \begin{aligned}
PE\left(\mathcal{H}_i,\mathcal{H}_j\right)=concat(z_{ij}^{\left(b,c\right)},z_{ij}^{\left(b,b\right)},z_{ij}^{\left(b,a\right)},z_{ij}^{\left(a,c\right)},...). 
\end{aligned}
\end{equation}
Such an operation allows the network to infer relative positioning of  \( \mathcal{H}_i\) and \(\mathcal{H}_j\) by comparing their temporal structures in feature space, which is also illustrated in Fig \ref{fig3}. Afterwards, ROR utilizes a fully connected layer, \(FC:\mathbb{R}^{C_3\times\left|\mathcal{P}_j\right|^2\ \ }\rightarrow\mathbb{R}^{\left|\mathcal{P}_j\right|!}, \) to optimize a Cross-Entropy loss:
\begin{equation}
    \begin{aligned}
        L_{ROR} = -\textbf{1}(\mathcal{P}_i | \mathcal{P}_j) \cdot \log \left(FC \left( PE(\mathcal{H}_i, \mathcal{H}_j) \right) \right)
    \end{aligned}
\end{equation}

\textbf{DuoCLR Loss.} The final loss function for DuoCLR combines CPC and ROR losses, with a weighting factor \(\alpha\) to balance their contributions:
\vspace{-5pt}
\begin{equation}
    \begin{aligned}
      L=\mathcal{L}_{CPC}+\alpha\mathcal{L}_{ROR}            
    \end{aligned}
\end{equation}
\vspace{-15pt}

The implementation of DuoCLR upon Eq. (12) leads to an expressive representation \(\mathcal{F}\) that is fully transferable to action segmentation and will be comprehensively evaluated in Section 4.


\section{Experiments}
\begin{figure}
    \centering
    \includegraphics[width=0.9\linewidth]{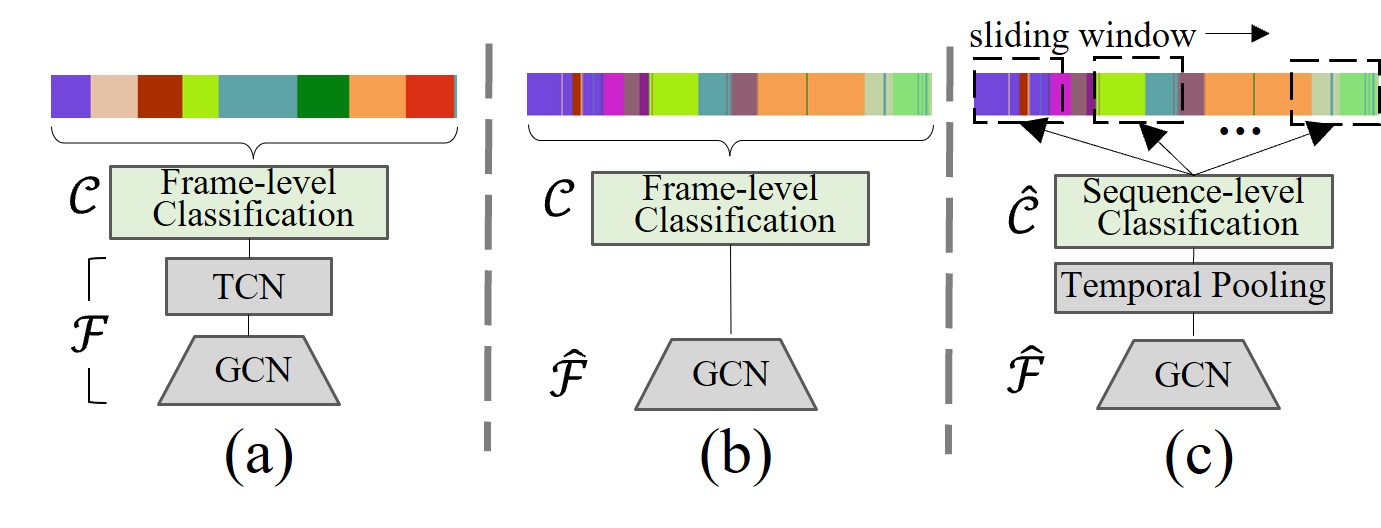}
    \vspace{-10pt}
    \caption{Transfer learning in skeleton-based human action segmentation. (a): Hierarchical structure introduced in Sec. 3.3. (b) and (c): Common structures used in existing works.  See more details in supplementary material.}
  \label{fig4}
\end{figure}
This section studies the application of DuoCLR in action segmentation. DuoCLR first involves a trimmed dataset for pretraining, followed by an evaluation phase where an untrimmed dataset is used to test the action segmentation performance. The study experimentally demonstrates the effectiveness of the proposed DuoCLR approach in comparison to state-of-the-art methods. 

\subsection{Implementations}

• \textbf{Datasets}. \textbf{Three trimmed} skeleton datasets are considered for pretraining in DuoCLR: NTU RGB+D (NTU) \cite{shahroudy2016ntu}, Toyota Smarthome (TS) \cite{das2019toyota} and Kinetics-400 (K4) \cite{kay2017kinetics}. \textbf{Three untrimmed} skeleton datasets are involved for evaluating the pre-trained models in the application of action segmentation, including PKU Multi-Modality Dataset (PKU) \cite{liu2017pku} (We use both versions: PKU1 and PKU2), Toyota Smarthome Untrimmed (TSU) \cite{dai2022toyota}, and Charades (CR) \cite{sigurdsson2016hollywood}. We report on mean Average Precision at IoU thresholds 0.1 (mAP$_{0.1}$) and 0.5 (mAP$_{0.5}$) \cite{yang2023lac}, Top1 Accuracy (Acc) \cite{yan2018spatial}, mean Intersection over Union (mIoU) \cite{lin2023actionlet}, and per-frame mAP (mAP) \cite{dai2022toyota} for  cross-view (CV) and cross-subject (CS) test cases of all untrimmed datasets.

• \textbf{Pretraining}. We adopt ST-GCN \cite{yan2018spatial} as the skeleton visual encoder and MS-TCN \cite{farha2019ms} as the skeleton temporal encoder. In the pretraining, we use the entire (both training and testing) set of a trimmed dataset to learn the feature extractor \(\mathcal{F}\). We adopt the same hyper-parameter values as in previous works \cite{guo2022contrastive, thoker2021skeleton, huang2023graph}
, i.e., \(C_3\) is set as 128 and \(\tau\)  in Eq. (1) as 0.07. The sizes of the two memory banks, \(\mathcal{N}^{\left(m\right)}\) and \(\mathcal{M}\), are set as 684 and 32,768, respectively. 
More details are available in supplementary material.

• \textbf{Evaluation}. We consider two popular downstream tasks: multi-class action segmentation \cite{liu2017pku, filtjens2022skeleton}  and multi-label (composite) action segmentation  \cite{yang2023lac, dai2022toyota}, in each of which we learn a frame-wise classification (frame annotating) layer \(\mathcal{C}\)   over the pre-learned \(\mathcal{F}\) (as illustrated in Fig. \ref{fig4}a). In multi-class segmentation, each skeleton frame is annotated with one action label. \(\mathcal{C}\) is composed of a  \(1\times1\) convolution and SoftMax layer. It is optimized by the multi-class Cross-Entropy loss function with one-hot encoded frame annotations of the evaluation dataset. In composite action segmentation, each skeleton frame could be labeled with multiple action labels (e.g., “Cooking” and “Stirring”). \(\mathcal{C}\) is composed of  a \( 1\times1\) convolution and Sigmoid layer and is optimized using the binary Cross-Entropy loss with multi-hot encoded frame annotations of the evaluation dataset. In both applications, we use SGD to update \(\mathcal{C}\) with Nesterov momentum 0.9 and learning rate 0.05.


\begin{table}[t]
    \centering

    \label{tab:multiclass}
    \fontsize{8}{9}\selectfont
    \begin{tabular}{@{}l|@{\hspace{2pt}}c@{\hspace{2pt}}c@{\hspace{2pt}}c@{\hspace{5pt}} p{0.45cm} p{0.45cm} p{0.45cm} p{0.45cm}}
        \toprule
        \textbf{Models} & \textbf{Method} & \textbf{Pretrain} & \textbf{Eval.} & \multicolumn{2}{c}{\textbf{CS\%}} & \multicolumn{2}{c}{\textbf{CV\%}} \\
        \cmidrule(lr){1-4} \cmidrule(lr){5-6} \cmidrule(lr){7-8} 
\rowcolor{gray!20} \textbf{NTU $\rightarrow$ PKU1:}& & & & \fontsize{6}{7}\selectfont mAP$_{.1}$ & \fontsize{6}{7}\selectfont mAP$_{.5}$ & \fontsize{6}{7}\selectfont mAP$_{.1}$ & \fontsize{6}{7}\selectfont mAP$_{.5}$ \\ 
        Baseline-I\cite{yan2018spatial} & Linear & NTU & PKU1 & \makebox[0.6cm][c]{56.0} & \makebox[0.6cm][c]{50.3} & \makebox[0.6cm][c]{56.7} & \makebox[0.6cm][c]{54.9} \\
        LAC \cite{yang2023lac} & Linear & Poetics & PKU1 & \makebox[0.6cm][c]{61.8} & \makebox[0.6cm][c]{-} & \makebox[0.6cm][c]{62.4} & \makebox[0.6cm][c]{-} \\
        \textbf{DuoCLR} & Linear & NTU & PKU1 & \makebox[0.6cm][c]{\underline{\textbf{85.2}}} & \makebox[0.6cm][c]{\underline{\textbf{74.3}}} & \makebox[0.6cm][c]{\underline{\textbf{84.7}}} & \makebox[0.6cm][c]{\underline{\textbf{77.5}}} \\
        Baseline-II \cite{filtjens2022skeleton} & F-T & -& PKU1 & \makebox[0.6cm][c]{89.6} & \makebox[0.6cm][c]{86.4} & \makebox[0.6cm][c]{92.1} & \makebox[0.6cm][c]{88.9} \\
        MS2L \cite{lin2020ms2l} & F-T & NTU & PKU1 & \makebox[0.6cm][c]{-} 
& \makebox[0.6cm][c]{50.9} 
& \makebox[0.6cm][c]{-} & \makebox[0.6cm][c]{-} \\
        CMD \cite{mao2022cmd} & F-T & NTU & PKU1 & \makebox[0.6cm][c]{-} 
& \makebox[0.6cm][c]{59.4} 
& \makebox[0.6cm][c]{-} & \makebox[0.6cm][c]{-} \\
        CrossSCLR \cite{li20213d} & F-T & NTU & PKU1 & \makebox[0.6cm][c]{-} 
& \makebox[0.6cm][c]{60.1} 
& \makebox[0.6cm][c]{-} & \makebox[0.6cm][c]{-} \\
        PCM3 \cite{zhang2023prompted} & F-T & NTU & PKU1 & \makebox[0.6cm][c]{-} 
& \makebox[0.6cm][c]{61.8} 
& \makebox[0.6cm][c]{-} & \makebox[0.6cm][c]{-} \\
        Hi-TRS \cite{chen2022hierarchically} & F-T & NTU & PKU1 & \makebox[0.6cm][c]{-} 
& \makebox[0.6cm][c]{63.5} 
& \makebox[0.6cm][c]{-} & \makebox[0.6cm][c]{-} \\
        USDRL \cite{weng2024usdrl}  & F-T & NTU & PKU1 & \makebox[0.6cm][c]{-} & \makebox[0.6cm][c]{75.7} & \makebox[0.6cm][c]{-} & \makebox[0.6cm][c]{-} \\
        LAC \cite{yang2023lac} & F-T & Poetics & PKU1 & \makebox[0.6cm][c]{92.6} & \makebox[0.6cm][c]{\textbf{90.6}} & \makebox[0.6cm][c]{94.6} & \makebox[0.6cm][c]{-} \\
        \textbf{DuoCLR} & F-T& NTU& PKU1& \makebox[0.6cm][c]{\underline{\textbf{94.4}}} & \makebox[0.6cm][c]{\underline{90.1}} & \makebox[0.6cm][c]{\underline{\textbf{96.8}}} & \makebox[0.6cm][c]{\underline{\textbf{94.8}}} \\

\rowcolor{gray!20} \textbf{NTU $\rightarrow$ PKU2:} & & & & \makebox[0.6cm][c]{mIoU} &  \makebox[0.6cm][c]{Acc} &  \makebox[0.6cm][c]{mIoU} & \makebox[0.6cm][c] {Acc} \\ 

        Baseline-I \cite{yan2018spatial}  & Linear & NTU & PKU2& \makebox[0.6cm][c]{30.7} & \makebox[0.6cm][c]{47.2} & \makebox[0.6cm][c]{26.4} & \makebox[0.6cm][c]{44.7} \\
        
        AimCLR \cite{guo2022contrastive}& Linear & Poetics & PKU2& \makebox[0.6cm][c]{-} & \makebox[0.6cm][c]{-} & \makebox[0.6cm][c]{15.7} & \makebox[0.6cm][c]{39.8} \\

        ActCLR \cite{lin2023actionlet} & Linear& NTU & PKU2 & \makebox[0.6cm][c]{-} & \makebox[0.6cm][c]{-} & \makebox[0.6cm][c]{21.4} & \makebox[0.6cm][c]{51.3} \\

        \textbf{DuoCLR} & Linear & NTU & PKU2 & \makebox[0.6cm][c]{\underline{\textbf{54.5}} } & \makebox[0.6cm][c]{\underline{\textbf{73.9}} } & \makebox[0.6cm][c]{\underline{\textbf{50.9}} } & \makebox[0.6cm][c]{\underline{\textbf{68.8}}} \\
     
        Baseline-II \cite{filtjens2022skeleton}  & F-T & -& PKU2 & \makebox[0.6cm][c]{50.9} & \makebox[0.6cm][c]{69.3} & \makebox[0.6cm][c]{45.6} & \makebox[0.6cm][c]{66.0} \\
        
       Hi-TRS \cite{chen2022hierarchically} & F-T & NTU & PKU2 & \makebox[0.6cm][c]{-} & \makebox[0.6cm][c]{-} & \makebox[0.6cm][c]{-} & \makebox[0.6cm][c]{55.0} \\
       
       LaSA \cite{ji2025language} & F-T & -& PKU2 & \makebox[0.6cm][c]{-} & \makebox[0.6cm][c]{73.4} & \makebox[0.6cm][c]{-} & \makebox[0.6cm][c]{69.4} \\
       
       \textbf{DuoCLR} & F-T & NTU & PKU2 & \makebox[0.6cm][c]{\underline{\textbf{66.1}} }& \makebox[0.6cm][c]{\underline{\textbf{82.6}}} & \makebox[0.6cm][c]{\underline{\textbf{56.3}}} & \makebox[0.6cm][c]{\underline{\textbf{79.2}}} \\
        \bottomrule
    \end{tabular}
\vspace{-10pt}
\caption{Experimental comparison in the multi-class skeleton-based action segmentation task. Evaluation methods include Linear Evaluation and Fine-tune (F-T) Evaluation.}
\label{table1}
\end{table}

\begin{table}[t]
    \centering
    \label{tab:composite}
    \fontsize{8}{9}\selectfont
    
    \begin{tabular}{l@{} c c p{0.5cm} p{0.5cm} c@{} c@{}}
        \toprule
        \multirow{2}{*}{Models} & \multirow{2}{*}{Methods}  &\multirow{2}{*}{Modality}& \multicolumn{2}{c}{\centering{\textbf{TS $\rightarrow$ TSU}}} & \multicolumn{2}{c}{\textbf{K4 $\rightarrow$ CR}} \\
        &  && CS(\%) & CV(\%) & \multicolumn{2}{c}{\centering mAP(\%)} \\
\midrule
 MS-TCT \cite{dai2022ms} &  F-T&RGB& 33.7& \centering{-}& \multicolumn{2}{c}{25.4}\\
        Baseline-I \cite{yan2018spatial} & Linear  &Skeleton& 18.2 & 11.4 & \multicolumn{2}{c}{6.1} \\
        LAC \cite{yang2023lac} & Linear  &Skeleton& 20.8 & 18.3 &  \multicolumn{2}{c}{14.3} \\
        \textbf{DuoCLR} & Linear  &Skeleton& 24.1 & 19.8 & \multicolumn{2}{c}{19.1} \\
        Baseline-II \cite{filtjens2022skeleton} & F-T  &Skeleton& 28.3 & 20.3 & \multicolumn{2}{c}{21.8} \\
        LAC \cite{yang2023lac} & F-T  &Skeleton& \textbf{36.8} & 23.1 &   \multicolumn{2}{c}{28.0} \\
        \textbf{DuoCLR} & F-T  &Skeleton& 35.1 & \textbf{26.3} & \multicolumn{2}{c}{\textbf{33.9}} \\
        \bottomrule
    \end{tabular}
\vspace{-10pt}
\caption{Experimental comparison in the composite skeleton-based action segmentation task.}
\label{table2}
\end{table}

\subsection{Results} 
In the multi-class action segmentation task, we use NTU RGB+D for pretraining while using two datasets, PKU1 and PKU2, as the target for evaluation. A similar setting is considered in composite action segmentation where the trimmed TS and K4 are used as the pretraining domain while TSU and CR are used for evaluation. As with previous works, we use Linear Evaluation \cite{li20213d}, where \(\mathcal{F}\) is fixed while training \(\mathcal{C}\), and Fine-tune (F-T) Evaluation \cite{li20213d}, where \( \mathcal{F}\) and \(\mathcal{C}\) are trained jointly, to evaluate our approach. We report the experimental results in Table \ref{table1} and Table \ref{table2}. 

•\textbf{ Comparison to State-of-the-art Methods.} We first compare the proposed DuoCLR to recent state-of-the-art methods\cite{yang2023lac, yang2021unik,dong2023hierarchical, duan2023skeletr, lin2020ms2l}  that involve a similar “trimmed-to-untrimmed” learning scenario. Experimental results demonstrate that DuoCLR outperforms most state-of-the-art comparatives across both multi-class (Table \ref{table1}) and composite (Table \ref{table2}) action segmentation tasks. We conclude that, since prior methods are limited in exploiting sequence-independent representations, the resultant feature space preserves a low transferability to action segmentation. In contrast, our model benefits from two efficient surrogate tasks, leading to better performance.

\vspace{-3pt}
• \textbf{Linear and Fine-tune Evaluation. }A vanilla supervised pretraining paradigm involves learning a feature space \(\mathcal{F} \)via a conventional end-to-end action recognition model \cite{yan2018spatial} using trimmed skeleton videos without data augmentation. We treat such a paradigm as a baseline model (Baseline-I in Table \ref{table1} and Table \ref{table2}) while making comparison with DuoCLR in Linear Evaluation. Experimental results demonstrate that our model significantly outperforms Baseline-I,   which validates the effectiveness of the proposed DuoCLR approach in learning expressive action representations for the action segmentation tasks. We also compare DuoCLR with Baseline-II, an action segmentation model \cite{filtjens2022skeleton}  that can be trained end-to-end on untrimmed datasets. DuoCLR, when fine-tuned, achieves substantial performance improvements over Baseline-II, as shown in Table \ref{table1} and Table \ref{table2} for both multi-class and composite tasks. This outcome highlights the advantage of incorporating DuoCLR’s pre-trained features, which enhance the model’s adaptability and performance in segmentation. 

\begin{table}[t]
    \centering
    \resizebox{0.45\textwidth}{!}{
 
    \begin{tabular}{llcccc}
        \toprule
        \multirow{2}{*}{Models} & \multirow{2}{*}{Methods} & \multicolumn{2}{c}{5\% of PKU1} & \multicolumn{2}{c}{10\% of PKU1} \\
        & & CS(\%) & CV(\%) & CS(\%) & CV(\%) \\
        \midrule
        Baseline-I & Linear & 33.9 & 36.0 & 41.6 & 42.7 \\
        DuoCLR & Linear & 58.4 & 41.3 & 70.0 & 72.8 \\
        Baseline II & Fine-Tune & 47.4 & 49.5 & 56.4 & 58.1 \\
 LAC\cite{yang2023lac} & Fine-Tune& 73.9& 75.4& 79.8&81.1\\
        DuoCLR & Fine-Tune & \textbf{74.6}& \textbf{77.5}& \textbf{81.9}& \textbf{84.2}\\
    
        \bottomrule
    \end{tabular}
    }
    \vspace{-10pt}
    \caption{Semi-supervised evaluation (mAP$_{0.1}$) with a low proportion of samples from the target domain.}
\label{table3}
\end{table}

\begin{table}[t]
\centering
\fontsize{8}{9}\selectfont
\begin{tabular}{l c c c c}
\hline
Model & PKU1 & PKU2 & TSU & Charade \\
DuoCLR & 63\% & 59\% & 55\% & 42\% \\
\hline
\end{tabular}
\vspace{-10pt}
\caption{The ratio of labeled data is saved while reaching the same performance with the vanilla model that was trained on 100\% data.}
\label{table4}
\end{table}
\vspace{-2pt}
\textbf{• Semi-supervised Evaluation} The last experiment examines the transferability of our model while using fewer data samples in evaluation. Specifically, the experiment varies the ratios (5\% and 10\%) of samples from the target domain involved in ransfer learning (i.e., Linear and Fine-tune Evaluation). Experimental results in Table \ref{table3} demonstrate that the model’s performance tends to increase monotonically along with the ratio of samples from the target domain. Table \ref{table4} suggests that DuoCLR can save up to 63\% labeled data to reach the compatible performance compared  to full-supervised learning. 
\begin{table}
\centering
\fontsize{8}{9}\selectfont
\begin{tabular}{l c c c}
\midrule

\textbf{NTU to PKU1}& Method& CS (\%) & CV (\%)\\
\midrule
Baseline-I& Linear & 56.0 & 60.3 \\
CPC & Linear & 83.9 & 85.1 \\
ROR & Linear & 76.7 & 79.5 \\
CPC+ROR (DuoCLR) & Linear & \textbf{85.2}& \textbf{84.7}\\
\midrule

\end{tabular}

\vspace{-10pt}
\caption{Ablation study of two surrogate tasks: CPC (Cross Permutation Contrasting), ROR (Relative Order Reasoning), and combined DuoCLR. Results are reported as mAP$_{0.1}$.}
\label{table5}
\end{table}

\begin{table}[t]
\centering
\fontsize{8}{9}\selectfont
{
\begin{tabular}{l |cc>{\centering\arraybackslash}p{0.7cm}>{\centering\arraybackslash}p{0.7cm}}
\midrule
\textbf{PKU1 to PKU2}& Methods& Pretraining& CS(\%)& CV(\%)\\
\midrule
Baseline-II & Fine-Tune & - & 69.3 & 66.0 \\
DuoCLR & Linear & Trimmed PKU1 & \textbf{78.1} & 76.0 \\
DuoCLR & Fine-Tune & Trimmed PKU1 & 78.0 & \textbf{77.6} \\
\hline
\end{tabular}
}
\vspace{-10pt}
\caption{Experimental results (Acc) of pretraining using untrimmed datasets.}
\label{table6}
\end{table}

\begin{figure}
    \centering
    \includegraphics[width=1\linewidth]{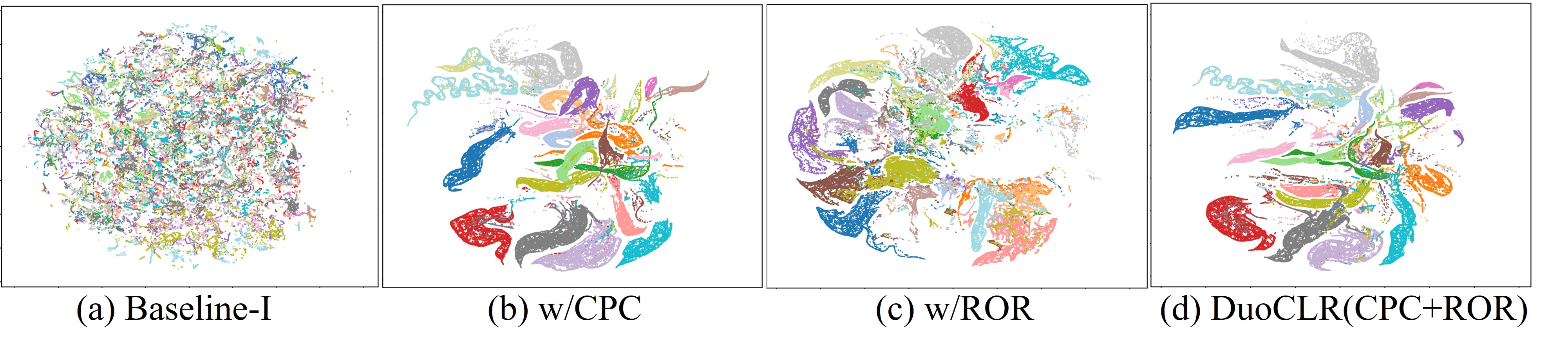}
\vspace{-20pt}
    \caption{Feature clusters visualization by t-SNE \cite{van2008visualizing}}
 \label{fig5}
\end{figure}
\subsection{Ablation studies}
• \textbf{Effectiveness of Surrogate Tasks.}  To assess the independent and combined effectiveness of these tasks, we conduct experiments using each task individually and in combination, pretraining on NTU and evaluating on PKU1. First, either surrogate task demonstrates a significant performance improvement compared to the vanilla supervised pretraining paradigm (Baseline-I) as summarized in Table \ref{table5}. In Fig. \ref{fig5}b and Fig. \ref{fig5}c, CPC and ROR demonstrate their independent effectiveness at learning more separate    feature clusters compared to Baseline-I  (Fig. \ref{fig5}a). Second, CPC achieves superior effectiveness over ROR likely because it directly enforces instance-level similarity, which is critical for clustering. Last, DuoCLR achieves its best learning result when combining the two tasks, i.e. 85.2\% and 84.7\% on the benchmarks of CS and CV, meanwhile demonstrating the best separability in feature clustering (Fig. \ref{fig5}d).

• \textbf{Pretraining with Untrimmed Datasets.} It is also effective to apply DuoCLR to an untrimmed dataset for representation learning. In this experiment, we consider PKU1 as the pretraining dataset where we trim each sequence into separate single-action sequences using their associated frame annotations. By applying “Shuffle and Warp” to these trimmed sequences, DuoCLR learns a feature extractor, \(\mathcal{F}\), which is then evaluated on the original untrimmed PKU2 dataset. As shown in Table \ref{table6}, DuoCLR outperforms the vanilla end-to-end supervised model (Baseline-II) in both linear and fine-tune evaluation protocols, showcasing the benefit of pretraining.

• \textbf{Effectiveness of Data Augmentation.}  We learn a new model (DuoCLR-I) by incorporating two classical skeleton augmentations, “Shear” and “Crop” \cite{li20213d}, with NTU used for pretraining and PKU1 for evaluation. As shown in Table \ref{table7}, DuoCLR-I achieves an accuracy of 58.3\%, suggesting that traditional augmentations alone provide limited transferability for action segmentation tasks. Our "Shuffle and Warp" method, however, is scalable to previous augmentations as we observe an orthogonal performance increase (1.3\% over regular DuoCLR in PKU1) with the model DuoCLR-III after combining two types of augmentation. Additionally, we perform an ablation study on the roles of "Shuffle" and "Warp" individually. The DuoCLR-II variant, which uses only "Shuffle" (Eq. (2)), achieves an accuracy of 73.8\%. When compared to DuoCLR, which uses both "Shuffle" and "Warp", the results demonstrate the importance of "Warp" which reduces irrelevant variations.
\begin{table}[t]
\centering
\resizebox{0.48\textwidth}{!}{
\begin{tabular}{l | c c c c | c}
\midrule
\textbf{NTU to PKU1} & Shear, Crop & Shuffle & Warp & Supervised & CS\% \\
\midrule
DuoCLR &   & $\surd$ & $\surd$ & $\surd$ & 85.2\\
DuoCLR-I & $\surd$ &   &   & $\surd$ & 58.3 \\
DuoCLR-II &   & $\surd$ &   & $\surd$ & 73.8 \\
DuoCLR-unsup & $\surd$ & $\surd$ & $\surd$ &   & 81.6 \\
DuoCLR-III & $\surd$ & $\surd$ & $\surd$ & $\surd$ & \textbf{87.1} \\
\hline
\end{tabular}
}
\vspace{-10pt}
\caption{Experimental comparison of DuoCLR and its variants using different skeleton data augmentation strategies.} 
\label{table7}
\end{table}

\begin{table}[t]
\fontsize{8}{9}\selectfont
\centering
\begin{tabular}{l @{}c >{\centering\arraybackslash}p{0.7cm} >{\centering\arraybackslash}p{0.7cm} >{\centering\arraybackslash}p{1.1cm} >{\centering\arraybackslash}p{0.4cm}}
\midrule
    Encoder& Classifier & Method & Params & Previous& Ours\\
\midrule
    GCN&Sequence-wise \(\hat{\mathcal{C}}\)& F-T & \centering{837K}& \centering{63.5 \cite{chen2022hierarchically}}& \textbf{77.6}  \\
  GCN& Frame-wise \(\mathcal{C}\)&  Linear & \centering{13.3K}& \centering{61.8 \cite{yang2023lac}}& \textbf{81.9}  \\
    GCN+TCN&Frame-wise \(\mathcal{C}\)&   Linear & \centering{13.3K}& \centering{-} & \textbf{85.2}  \\
\hline

\end{tabular}
\vspace{-10pt}
\caption{Experimental comparison between two types of classifiers. For \(\hat{\mathcal{C}}\), the sliding window size is set as 300. }
\label{table8}
\end{table}

• \textbf{Pretraining with Unlabeled Datasets.} DuoCLR also supports unsupervised pretraining. For this approach, labeled proxies are created using sequence subscripts within each data batch. To learn an unsupervised DuoCLR model (DuoCLR-unsup, Table \ref{table7}), we complement the augmentation with “Shear” and “Crop” to compensate for the lack of explicit labels. As seen in Table \ref{table7}, regular DuoCLR outperforms the unsupervised variant (81.6\% vs. 85.8\% accuracy). This performance gap is likely due to the challenge of accurately identifying negative pairs in an unsupervised setting, which can lead to noise to the learning process.

• {\textbf{DuoCLR is general to different network structures.}  Fig. \ref{fig4} illustrates two network structures commonly used in skeleton action segmentation: 1) Siding-window-free structure (Fig. \ref{fig4}b) composed of a GCN encoder \(\hat{\mathcal{F}}\) and a frame-wise classifier \(\mathcal{C}\); 2) sliding window based structure (Fig. \ref{fig4}c) composed of a  \(\hat{\mathcal{F}}\) and a sequence-wise classifier  \(\hat{\mathcal{C}}\) while using sliding windows to pre-segment input sequences. Fig. \ref{fig4}a illustrates the network proposed in Sec. 3.3. For evaluation, we train three DuoCLR models with different structures and conclude the experimental results in Table \ref{table8}). It demonstrates that DuoCLR is effective for different structures while suppassing previous works with clear margins.

•\textbf{ Effectiveness of Action Granularity.} In this study, we treat the action granularity (\(\mathcal{G}\)) of action permutations as a hyperparameter and learn how it affects optimizations of DuoCLR. In practice, setting a high value of \(\mathcal{G}\) will introduce diverse action contexts, however, it can also lead to high computational complexity, e.g., \(o(\mathcal{G})!\) for ROR while the size of the FC layer grows. In experiments, we increase \(\mathcal{G}\) from 1 to 5 while preserving other implementations parameters as defined in Table \ref{table1}. We illustrate confusion matrices (showing how ROR is good at mapping permutations) and similarity matrices (showing how CPC is good at learning consistent features) in Fig. \ref{fig6}, where Acc values evaluate the transferability of the resultant (CPC+ROR) network on PKU 2.  While \(\mathcal{G}=1\), it degrades to a vanilla contrastive learning model \cite{guo2022contrastive}. While \(\mathcal{G}\) increases, each model is capable of learning at different action granularity and and the best model is achieved when \( \mathcal{G}\) is around 4.

\begin{figure}
    \centering
    \includegraphics[width=0.9\linewidth]{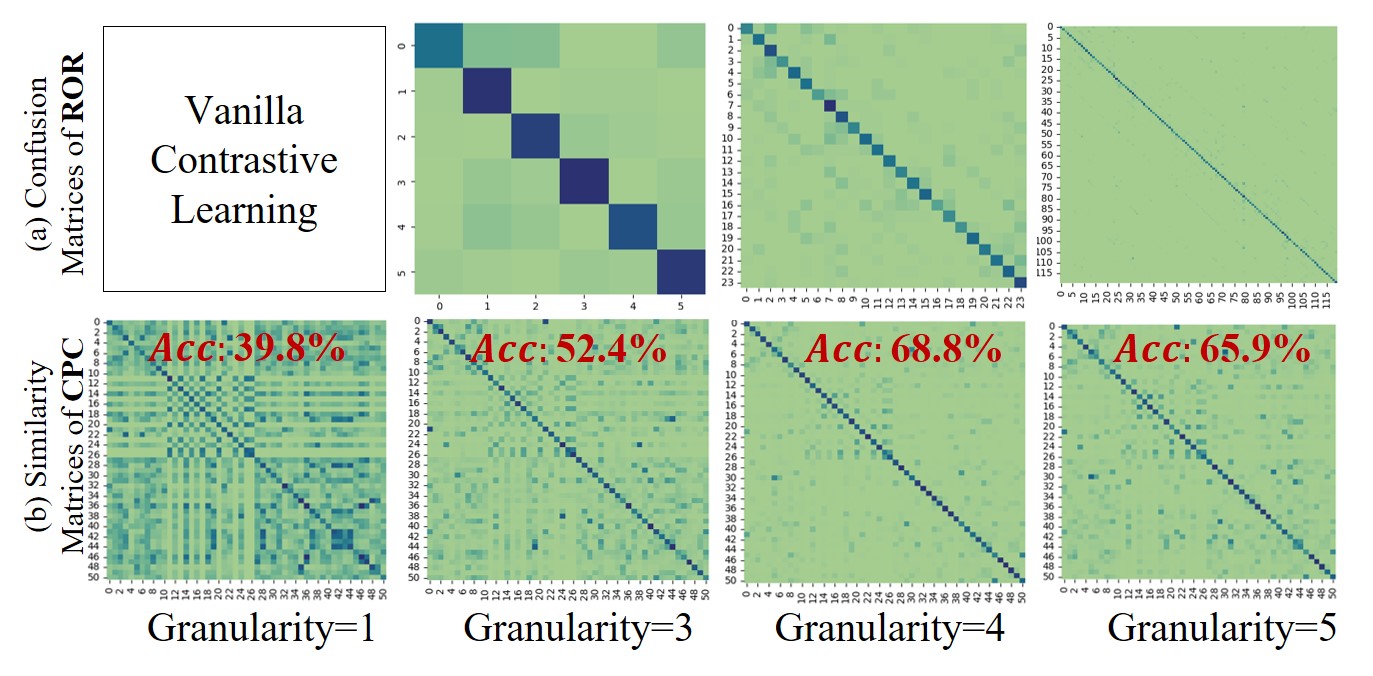}
 \vspace{-10pt}
    \caption{The confusion matrices obtained by ROR on different action granularities.}
 \label{fig6}
\end{figure}

\section{Conclusions}

This paper introduces a dual-surrogate contrastive learning (DuoCLR) framework while exploiting fine-grained extraction on action representations for skeleton data-based human action segmentation. It applies “Shuffle and Warp” to generate various multi-action permutations where two surrogate tasks, Cross Permutation Contrasting and Relative Order Reasoning, learn permutation-invariant intra-class similarities and permutation-aware inter-class contexts. The learned action representations can be represented by a feature extractor which can be entirely transferred into action segmentation. Experimental results demonstrate that 1) DuoCLR enables training with both unlabeled and labeled trimmed skeleton videos; 2) DuoCLR outperforms state-of-the-art methods in multiple tasks; 3) DuoCLR can save 42\%-63\% frame-level labeled data to reach the compatible fully-supervised performance in action segmentation; 4) DuoCLR supports learning with RGB videos that can lead to better action segmentation performance compared to previous RGB based approaches. 
\section{Acknowledgments}

This research was supported in part by MITACS Accelerate and NSERC Discovery grants. The authors would like to express their sincere gratitude to Dr. Pierre Payeur for his invaluable contributions to this work. Dr. Payeur provided consistent supervision throughout the research and played a key role in project administration and funding acquisition. His thoughtful input during the review and editing process was instrumental to the successful completion of this work. The authors also gratefully acknowledge the collaboration of the Sensing and Machine Vision for Automation and Robotic Intelligence (SMART) laboratory and Spectronix Inc.
{
    \small
    \bibliographystyle{ieeenat_fullname}
    \bibliography{main}
}


\end{document}